\documentclass[sigconf]{acmart}
\AtBeginDocument{%
  }
\usepackage{multirow}
\usepackage{amsmath}
\usepackage{booktabs}
\usepackage{amsfonts}

\copyrightyear{2026}
\acmYear{2026}
\setcopyright{cc}
\setcctype{by}
\acmConference[KDD '26]{Proceedings of the 32nd ACM SIGKDD Conference on Knowledge Discovery and Data Mining V.2}{August 09--13, 2026}{Jeju Island, Republic of Korea}
\acmBooktitle{Proceedings of the 32nd ACM SIGKDD Conference on Knowledge Discovery and Data Mining V.2 (KDD '26), August 09--13, 2026, Jeju Island, Republic of Korea}
\acmDOI{10.1145/3770855.3818453}
\acmISBN{979-8-4007-2259-2/2026/08}

\begin{document}

\title{NSR-Boost: A Neuro-Symbolic Residual Boosting Framework for Industrial Legacy Models}


\author{Ziming Dai}
\authornote{Both authors contributed equally to this research.}
\affiliation{%
  \institution{Tianjin University}
  \city{Tianjin}
  \country{China}}
\email{phoenixdai@tju.edu.cn}

\author{Dabiao Ma}
\authornotemark[1]
\affiliation{%
  \institution{Qfin Holdings, Inc.}
  \city{Beijing}
  \country{China}
}
\email{madabiao-jk@qifu.com}

\author{Jinle Tong}
\affiliation{%
  \institution{Qfin Holdings, Inc.}
  \city{Beijing}
  \country{China}
}
\email{lancertong@live.com}

\author{Mengyuan Han}
\affiliation{%
  \institution{Qfin Holdings, Inc.}
  \city{Beijing}
  \country{China}
}
\email{hanmengyuan-jk@qifu.com}

\author{Jian Yang}
\affiliation{%
  \institution{Qfin Holdings, Inc.}
  \city{Beijing}
  \country{China}
}
\email{wangye3-jk@qifu.com}

\author{Hongtao Liu}
\affiliation{%
  \institution{Qfin Holdings, Inc.}
  \city{Beijing}
  \country{China}
}
\email{htliu@tju.edu.cn}

\author{Haojun Fei}
\affiliation{%
  \institution{Qfin Holdings, Inc.}
  \city{Beijing}
  \country{China}
}
\email{zhangchulan-jk@qifu.com}

\author{Qing Yang}
\authornote{Corresponding author.}
\affiliation{%
  \institution{Qfin Holdings, Inc.}
  \city{Beijing}
  \country{China}
}
\email{yangqing3-jk@qifu.com}

\renewcommand{\shortauthors}{Ziming Dai et al.}

\begin{abstract}
  Although the Gradient Boosted Decision Trees (GBDTs) dominate industrial tabular applications, upgrading legacy models in high-concurrency production environments still faces prohibitive retraining costs and systemic risks. To address this problem, we present NSR-Boost, a neuro-symbolic residual boosting framework designed specifically for industrial scenarios. Its core advantage lies in being ``non-intrusive''. It treats the legacy model as a frozen model and performs targeted repairs on ``hard regions'' where predictions fail. The framework comprises three key stages: First, finding hard regions through residuals, then generating interpretable experts by generating symbolic code structures using Large Language Model (LLM) and fine-tuning parameters using Bayesian optimization, and finally dynamically integrating experts with legacy model output through a lightweight aggregator. Experimental results demonstrate that the framework significantly outperforms state-of-the-art (SOTA) baselines across six public datasets and one private dataset. More importantly, we report the successful deployment of NSR-Boost within the core financial risk control system of Qfin Holdings, where empirical results on real-world online traffic exhibit superior performance improvements and a significant reduction in the bad rate. In conclusion, it effectively captures long-tail risks missed by traditional models and offers a safe, low-cost evolutionary paradigm for industry.
\end{abstract}



\maketitle

\section{Introduction}
In the fields of computer vision and natural language processing, deep learning has demonstrated exceptional problem-solving capabilities~\cite{noor2025survey}. However, in industrial settings, especially in core business scenarios, tabular data remains the foundational pillar for supporting business operations~\cite{li2025treexformer}. In these high-risk, low-latency application environments, Gradient Boosted Decision Trees (GBDTs, such as XGBoost~\cite{chen2016xgb} and LightGBM~\cite{ke2017lightgbm}) consistently dominate as the de facto industrial standard due to their outstanding interpretability, training efficiency, and inference speed.

Nevertheless, standard GBDT models are gradually approaching the bottleneck of their representation capability. The existing improvement approaches primarily fall into two categories: One focuses on proposing novel model architectures (e.g., CatBoost~\cite{prokhorenkova2018catboost} and TabNet~\cite{arik2021tabnet}) to enhance overall performance; the other concentrates on feature engineering (e.g., OpenFE~\cite{zhang2023openfe}), attempting to assist GBDT in capturing non-linear relationships through automated feature generation. However, both categories face a critical common challenge in industrial deployment. Since industrial legacy models typically have undergone prolonged expert tuning and system integration, either completely replacing the model architecture or introducing large-scale new features requires refactoring the entire engineering pipeline. This not only incurs high maintenance costs and potential system risks but is also often unacceptable to many enterprises that prioritize stability. Therefore, there is an urgent need in the industry for a non-intrusive framework that can achieve incremental improvements on legacy models without disrupting the existing deployment ecosystem.

The emergence of Large Language Models (LLMs) provides a brand new solution perspective for this challenge. LLMs have not only demonstrated remarkable capabilities in code generation, as evidenced by AlphaEvolve~\cite{novikov2025alphaevolve} and LLM-SR~\cite{shojaeellm}, but more importantly, they possess prior knowledge regarding semantic information processing. This capability effectively compensates for the limitations of numerical GBDT models in logical reasoning. Inspired by this, we reframe complex feature interaction and residual correction tasks as a code generation problem and propose NSR-Boost (\textbf{N}euro-\textbf{S}ymbolic \textbf{R}esidual \textbf{Boost}ing). This is a novel, model-agnostic framework designed to patch industrial legacy models rather than reconstructing them from scratch.

Unlike existing methods that treat the LLM as a black-box predictor, NSR-Boost repositions the LLM as a symbolic code generator. We regard the existing GBDT model as a frozen model and utilize LLM to generate expert functions specifically for regions where the legacy model performs poorly. Our framework guides the LLM to uncover complex non-linear interaction logic that is difficult for GBDTs to capture. Crucially, the final output of NSR-Boost consists of executable Python code. This implies that during the actual deployment phase, we completely eliminate the inference cost of LLMs while preserving full logical interpretability.

Specifically, NSR-Boost introduces a bi-level optimization strategy to ensure its stability. In the outer loop, the LLM continuously refines the structure of Symbolic Experts based on ``Negative/Positive Constraints''. In the inner loop, the LLM autonomously generates the Bayesian search space and executes the optimization scripts to fine-tune the parameters of the generated functions, which effectively addresses the challenge of simultaneous structure and parameter search in symbolic regression. Finally, a context-aware aggregator dynamically integrates the legacy model with the generated experts to ensure a ``safe improvement'' over the legacy model. Our main contributions are summarized as follows:
\begin{itemize}
    \item We propose NSR-Boost, the first neuro-symbolic framework that leverages LLM-generated code to improve performance while keeping the legacy model frozen, thereby achieving low-latency inference and full logical interpretability.
    
    \item We design a bi-level optimization mechanism that explicitly decouples symbolic evolution from parameter tuning. By combining structure search based on LLMs with gradient-free Bayesian fine-tuning, our framework effectively overcomes the non-differentiability bottleneck of symbolic regression and enables the precise refinement of coarse-grained distribution boundaries.

    \item Extensive experiments demonstrate that NSR-Boost not only outperforms state-of-the-art (SOTA) baselines on six public benchmarks and one private dataset, but also significantly enhances legacy model performance on real-world financial data from Qfin Holdings.
\end{itemize}

\section{Related Work}
\textbf{Tabular Representation Learning.} For a long time, GBDTs (such as XGBoost~\cite{chen2016xgb} and LightGBM~\cite{ke2017lightgbm}) have established their de facto standard position in industrial tabular modeling due to their robustness against irregular data and efficient inference capabilities. In recent years, researchers have attempted to introduce neural networks into this domain. For example, TabNet uses a sequential attention mechanism for feature selection~\cite{arik2021tabnet}, while FT-Transformer introduces self-attention mechanisms to capture complex interactions~\cite{gorishniy2021revisiting}. Although these deep architectures have achieved significant progress on certain academic benchmarks, their industrial deployment still faces formidable challenges.

First, compared to GBDTs, Transformer-based models are typically associated with high inference latency, making it difficult to meet the response requirements of online services. Second, deep models lack inductive bias for tabular data and are extremely sensitive to hyperparameters. Consequently, they often fail to consistently outperform expert-tuned GBDTs in noisy industrial scenarios~\cite{shwartz2022tabular}. Given this, NSR-Boost does not seek a high-cost model replacement. Instead, it treats the GBDT as a frozen, efficient model and achieves non-intrusive performance enhancement through an external neuro-symbolic mechanism.

\noindent \textbf{Automated Feature Engineering.} Automated Feature Engineering (AFE) aims to assist GBDT models by mining high-order feature interactions. Early AFE methods primarily relied on heuristic search or predefined operator transformations. For example, AutoFeat~\cite{horn2019autofeat} and OpenFE~\cite{zhang2023openfe} generate candidate features by iteratively applying mathematical operations and use statistical metrics to filter for effective features. However, these methods are often constrained by a vast search space and struggle to leverage the semantic information embedded in feature names.

With the advent of LLMs, researchers have begun to utilize the semantic priors of LLMs to guide feature generation. CAAFE~\cite{hollmann2023large} and FeatLLM~\cite{han2024large} pioneer the approach of prompting LLMs to generate physically meaningful feature formulas based on feature descriptions. Recent works have further advanced this field by introducing evolutionary optimization strategies and reasoning-type exploration mechanisms, respectively, to enhance search efficiency~\cite{abhyankar2025llm,han2025tabular}. However, all the aforementioned methods suffer from a fatal drawback regarding industrial deployment. The static new features they generate force changes to the input data schema. This not only needs the full retraining of downstream GBDT models but also requires the refactoring of online feature extraction pipelines, incurring substantial maintenance costs and deployment risks. In contrast, NSR-Boost does not alter the input feature space, thereby avoiding the aforementioned risks.

\noindent \textbf{Code Generation.} Modeling regression and discovery tasks as code generation has emerged as a research hotspot in recent years. The pioneering works of FunSearch~\cite{romera2024mathematical} and AlphaEvolve~\cite{novikov2025alphaevolve} have demonstrated that LLMs possess the potential to discover novel algorithms and mathematical knowledge via evolutionary strategies. Building upon this foundation, LLM-SR~\cite{shojaeellm} and its improved version, DrSR~\cite{wang2025drsr}, introduce this paradigm into symbolic regression. They utilize LLMs to directly generate Python programs to fit data and uncover underlying scientific laws. To further enhance performance, Grayeli \textit{et al.} propose a knowledge reuse mechanism based on a concept library~\cite{grayeli2024symbolic}, while LLM-Meta-SR explores meta-learning for evolutionary selection operators~\cite{zhang2025llm}.

However, the direct application of existing Symbolic Regression (SR) methods faces two primary challenges. First, SR typically aims to discover global formulas to fully replace the original model. In complex industrial scenarios, abandoning the GBDT model that has undergone long-term tuning implies extremely high deployment risks and system instability. Second, existing methods predominantly rely on gradient descent for parameter optimization~\cite{shojaeellm}. This dependence imposes a strict constraint on function differentiability, thereby rendering them incapable of handling discrete logic that characterizes business rules. In light of this, NSR-Boost only performs local residual repair, and introduces gradient-free Bayesian optimization, overcoming differentiability constraints and achieving high-precision fine-tuning for any complex logic.

\section{Notations and Problem Formulation}
\label{problem}
\begin{figure*}[t]
    \includegraphics[width=\textwidth]{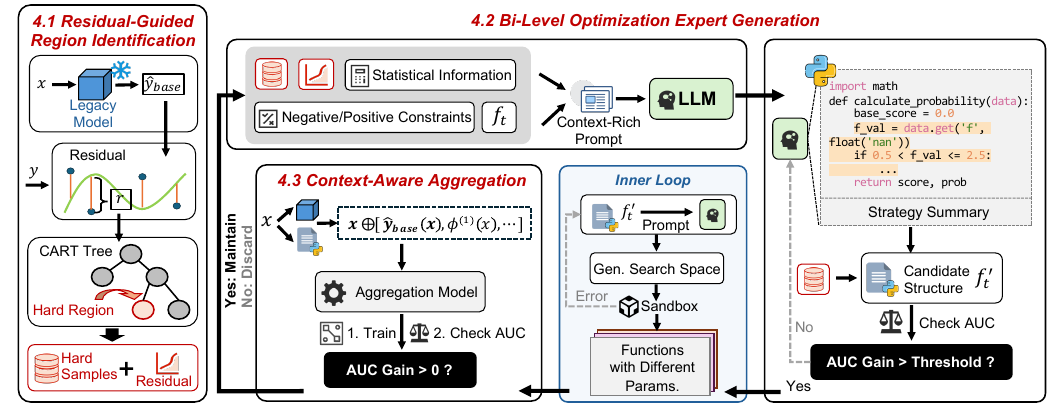}  
    \caption{The framework overview of NSR-Boost.}
    \label{fig:framework}
\end{figure*}

Consider a standard tabular classification (or regression) task defined on the dataset $\mathcal{D} = \{(\mathbf{x}_i, y_i)\}_{i=1}^N$. We assume the existence of a deployed legacy model $\mathcal{M}_{base}$ , with a prediction output $\hat{y}_{base}(\mathbf{x})$ for an input $\mathbf{x}$. The key constraint is that, to strictly adhere to industrial requirements for system stability, we treat $\mathcal{M}_{base}$ as a frozen model. Consequently, any modification to its parameters or structure is strictly prohibited throughout the optimization process.

The core objective of NSR-Boost is to learn an additive correction term $F$ without altering the parameters of $\mathcal{M}_{base}$, such that the generalization performance of the final hybrid model $\hat{y}_{final}(\mathbf{x}) = \hat{y}_{base}(\mathbf{x}) + F(\mathbf{x})$ is maximized on the full validation set $\mathcal{D}_{val}$.

A direct search for the optimal non-linear Symbolic Expert $F$ within the global space constitutes a highly complex combinatorial optimization problem. To reduce computational complexity and ensure interpretability, we decouple this global objective into a set of parallel local incremental learning sub-problems. Let $\mathbb{P} = \{\mathcal{D}_1, \dots, \mathcal{D}_K\}$ be a partition of the target feature space, where $\mathcal{D}_k \subset \mathcal{D}$ denotes a specific local feature subspace (the specific partitioning strategy is detailed in Section~\ref{sec:region_id}). Based on this decomposition, our optimization process is formalized into the following two hierarchical phases:

\textbf{P1. Local Subspace Optimization.}
For each local subspace $\mathcal{D}_{k}$, we aim to construct a Symbolic Expert $f^{(k)}(\cdot; S_k, \boldsymbol{\theta}_k)$. Our objective is to search for the optimal symbolic structure $S_k$ and parameters $\boldsymbol{\theta}_k$ such that the hybrid model composed of the legacy model and the expert minimizes the task-specific loss $\mathcal{L}$ strictly within this local region:
\begin{equation}
    \min_{S_k, \boldsymbol{\theta}_k} \sum_{(\mathbf{x}, y) \in \mathcal{D}_{k}} \mathcal{L}\left(y, \sigma(\hat{z}_{base}(\mathbf{x}) + f^{(k)}(\mathbf{x}; S_k, \boldsymbol{\theta}_k))\right),
\end{equation}
where $\hat{z}_{base}$ denotes the logit output of the legacy model, and $\sigma$ is the activation function.

\textbf{P2. Global Aggregation Optimization.}
Given the library of experts $\mathcal{F}=\{f^{(1)}, \dots, f^{(K)}\}$ generated in \textbf{P1}, the objective of the second phase is to learn a global aggregation strategy $\mathcal{G}$ to minimize the global loss on the full validation set $\mathcal{D}_{val}$, which constitutes the ultimate optimization objective of our system:
\begin{equation}
      \min_{\mathcal{G}} \sum_{(\mathbf{x}, y) \in \mathcal{D}_{val}} \mathcal{L}\left(y, \sigma(\hat{z}_{base}(\mathbf{x}) + \mathcal{G}(\mathbf{x}, \mathcal{F}))\right),
\end{equation}
where $\mathcal{G}$ is an abstract mapping responsible for coordinating the outputs of multiple local experts, ensuring that the model enhances overall generalization performance and robustness.

\section{Methodology}

In this section, we present the overall architecture of NSR-Boost, which is organized into three tightly coupled phases, as illustrated in Figure~\ref{fig:framework}. First, we analyze the residual distribution of the frozen model and employ a CART tree to identify the hard region $\mathcal{D}_h$ as the target for subsequent optimization. This ensures that computational resources are concentrated on the model's weakest points. Subsequently, we initiate $K$ independent chains, adopting a bi-level optimization strategy for expert code generation. Here, we introduce a Cascaded Validation mechanism to ensure that the selected experts contribute substantial metric improvements to the legacy model. Finally, during the inference phase, we deploy a lightweight context-aware aggregator to perform dynamic arbitration on the outputs of each expert, yielding the final prediction.

\subsection{Residual-Guided Region Identification}
\label{sec:region_id}

Before the generation stage is initiated, it is essential to precisely localize the target region for optimization. Unlike the traditional strategies that blindly fit the global distribution, we fully leverage the prior knowledge embedded in the legacy model. We quantify its performance deficits by calculating the residuals $r_i = y_i - \hat{y}_{base}(\mathbf{x}_i)$ on the training set. Samples exhibiting significant residuals are defined as ``Hard Samples'', representing bottleneck regions where an immediate enhancement in model performance is required.

To precisely map the distribution of these samples within the feature space and construct the partition $\mathbb{P}$ defined in Section~\ref{problem}, we train a shallow CART tree to fit the absolute residuals $|r_i|$~\cite{loh2011classification}. We don't need to be overly precise here, as we will refine the distribution boundaries for it in section~\ref{sec:bi_level}. Subsequently, we introduce a weighted scoring mechanism that calculates a priority score $C$ for each leaf node based on its cumulative error and sample coverage. Leaf nodes with scores exceeding a predefined threshold are identified as ``hard regions''. We define the union of samples covered by these selected nodes as the global hard region $\mathcal{D}_{h}$. Crucially, leveraging the inherent disjoint nature of CART tree leaves, $\mathcal{D}_{h}$ is naturally partitioned into multiple disjoint sub-regions, where each region serves as the specific feature subspace for the parallel chain. This strategy effectively balances error magnitude with sample size, guiding the LLM to focus on high-potential regions that offer the greatest leverage for maximizing model performance.

There are several motivations for introducing the CART tree. First, its computational overhead is negligible, imposing no significant training burden. Second, the explicit decision rules derived from the CART tree (e.g., \texttt{Age > 50}) can be embedded into the Prompt as a semantic prior. This distinct boundary information not only assists the LLM in generating highly targeted Symbolic Experts but also effectively mitigates noise interference from expert on samples in non-target regions. Finally, benefiting from the mutual exclusivity of CART tree leaf nodes, the derived regions are mathematically disjoint. This guarantees that the optimization processes across different chains remain non-interfering.

\subsection{Bi-Level Optimization Expert Generation}
\label{sec:bi_level}

This section details the core of NSR-Boost, aiming to solve the \textbf{P1}. Given the partition $\mathbb{P}$ established in Section~\ref{sec:region_id}, the framework orchestrates $K$ parallel evolutionary chains. Without loss of generality, this section focuses on the internal mechanism of a single chain dedicated to a specific local subspace $\mathcal{D}_k$.

We decouple the search for discrete symbolic structures from continuous parameter optimization into two nested loops. This strategy stems from the inherent discrete-continuous duality of SR tasks: identifying the optimal functional form is essentially a discrete combinatorial optimization problem, which aligns naturally with the logical reasoning strengths of LLMs~\cite{romera2024mathematical}; conversely, determining optimal constant coefficients constitutes a continuous numerical optimization problem demanding high precision, which typically exposes the limitations of token-based prediction in LLMs~\cite{schick2023toolformer}. This decoupling effectively mitigates the combinatorial explosion challenge associated with mixed search spaces.

\noindent \textbf{Outer Loop.} We formalize the outer loop as an LLM-driven, data-aware incremental optimization process. To avoid the randomness of LLM's generation from scratch and to regularize the output space, the system maintains a currently optimal ``Seed Function'' and guides the LLM to perform directed, incremental modifications based on the real data distribution.

The Seed Function is initialized as a constant bias term constrained by region rules (i.e., $f_0(x) = c$). Specifically, to achieve precise spatial decoupling of expert models in the feature space, we explicitly encode the region definition rules derived from the CART tree as a conditional statement. This mandates that the function is activated solely within the specific hard region from its very inception. Subsequently, building upon this restricted Seed Function, the system executes $T$ iterative cycles to progressively evolve complex residual correction logic.

In the $t$-th iteration, we construct a context-rich prompt $\mathcal{P}_t$, as detailed in Appendix~\ref{apx:prompt}. Its purpose is to guide the LLM to generate symbolic structures that are both meaningful and fully consistent with the data distribution. To bridge the gap between abstract logic and concrete numerical distributions, $\mathcal{P}_t$ integrates critical information across three dimensions. First, we incorporate the task description, incremental modification rules, and the current Seed Function $f_{t-1}$ alongside its performance metric $A_{t-1}$, thereby explicitly defining the starting point for optimization. Second, to enhance data awareness, we inject feature statistical information (e.g., IV and PSI) to highlight high-potential features and directly sample $N$ real instances from the sample region $\mathcal{D}_k$ as few-shot context. Crucially, we annotate these instances with fine-grained error types (e.g., False Negatives, High Uncertainty), compelling the LLM to explicitly attend to specific failure modes. Finally, we introduce a historical feedback loop by embedding summaries of previously failed and successful modifications as ``Negative/Positive Constraints'' within the prompt, effectively preventing the process from cycling through repetitive, invalid trial and error.

Based on $\mathcal{P}_t$, the LLM outputs a modified candidate structure $f'_t$ along with a textual summary of the modification intent. Subsequently, this structure enters an inner loop to complete parameter fine-tuning (detailed below) and evaluates its  effectiveness with the legacy model via a lightweight proxy model. To balance exploration and exploitation, we introduce a dynamic annealing threshold $\tau$ that decays over iterations. If and only if the fusion metric (Area Under the Curve, AUC) surpasses $A_{t-1} - \tau$, we promote $f'_t$ as the new Seed Function. This mechanism allows the algorithm to tolerate minor performance fluctuations in the early stage to expand the search space and avoid prematurely falling into local optima. Both successful and failure modification summaries will be placed in the prompt to enter the next loop.

\noindent \textbf{Inner Loop.} Once the outer loop determines the candidate structure $f'_t$, the inner loop is immediately initiated. Its objective is to efficiently and robustly determine the optimal continuous parameter $\theta$ through a dual iterative mechanism. 

To avoid the syntactic risks introduced by LLM generating complete code, at this stage, the system pre-integrates a standard Tree-structured Parzen Estimator (TPE) optimization framework. The LLM is strictly restricted to the role of a ``Configurator'', responsible solely for generating code snippets that define parameters and search spaces~\cite{maeureka}. The prompt is shown in Appendix~\ref{apx:prompt}. These configuration snippets are injected into a sandbox environment for attempted execution. Upon capturing any runtime exceptions, the error stack traces are converted into debugging reports and fed back to the LLM, triggering a targeted correction of the parameter definition logic\cite{wang2023voyager}. This ``Generation-Injection-Error-Correction'' closed loop iterates until the script is successfully launched, thereby ensuring the robustness of the automated workflow.

Upon the successful loading of the configuration, the system initiates the Bayesian optimization iteration. The TPE algorithm performs $M$ consecutive sampling and evaluation steps within the parameter space to minimize residual loss. To strike a balance between search efficiency and fitting precision, we employ an Incremental Freezing Strategy: Parameters inherited from the previous generation's seed function are frozen, while the search space is exclusively restricted to the coefficients of the structures added in the current round.

Crucially, to compensate for the discretization error induced by the depth constraints of the CART tree, a hard threshold of global conditional statements generated by the outer loop is explicitly incorporated into the hyperparameter search space during the first successful Bayesian optimization run. Via numerical iterative approximation, the optimizer is able to fine-tune these coarse-grained boundaries, for example, adjusting a split point from 50 to 48.5, thereby realizing a seamless transition from discrete symbolic rules to precise numerical boundaries.

\subsection{Context-Aware Aggregation}
\label{sec:global_assembly}

We obtain a library of $K$ local Symbolic Experts $\mathcal{F}=\{f^{(1)}, \dots, f^{(K)}\}$, where each expert $f^{(k)}$ has been optimized for its specific feature subspace $\mathcal{D}_k$ (addressing \textbf{P1}). While each expert effectively reduces residuals within its local region, a direct linear summation is often insufficient to address boundary smoothness or dynamically adapt to the complex global distribution of the legacy model. To solve the \textbf{P2} and instantiate the abstract aggregation strategy, this section proposes a context-aware aggregation strategy based on symbolic interaction vectors.

To enable the downstream aggregation model to fully comprehend the non-linear relationship between the Symbolic Experts and the legacy model, relying solely on the raw outputs of the experts is inadequate. Instead, for an arbitrary expert $f^{(k)}$ and an input sample $\mathbf{x}$, we construct an enhanced Symbolic Interaction Vector $\phi^{(k)}(\mathbf{x})$ to explicitly capture the raw score, the magnitude of residual correction, and the relative ratio:
\begin{equation}
    \phi^{(k)}(\mathbf{x}) = [ \ \underbrace{f^{(k)}(\mathbf{x})}_{\text{Score}}, \ \ \underbrace{(f^{(k)}(\mathbf{x}) - \hat{y}_{base}(\mathbf{x}))}_{\text{Residual } \delta^{(k)}}, \ \ \underbrace{\frac{f^{(k)}(\mathbf{x})}{\hat{y}_{base}(\mathbf{x}) + \epsilon}}_{\text{Ratio } \pi^{(k)}} \ ].
\end{equation}

Building upon this feature representation, we introduce a lightweight XGBoost $\mathcal{M}_{g}$ to serve as the global aggregation model. Distinct from traditional Stacking strategies that rely solely on prediction scores, we construct a context-aware input space by concatenating the raw business features $\mathbf{x}$ with the legacy model output and the interaction vectors of the currently active experts. This design empowers the Context-Aware Gating to go beyond merely utilizing numerical expert outputs; it allows for the dynamic arbitration of correction effectiveness based on the raw feature distribution, thereby facilitating robust decision-making under complex distributions. The final prediction output $\hat{y}_{final}(\mathbf{x})$ is defined as follows:
\begin{equation}
    \hat{y}_{final}(\mathbf{x}) = \mathcal{M}_{g}\left( \mathbf{x} \oplus \left[ \hat{y}_{base}(\mathbf{x}), \phi^{(1)}(\mathbf{x}), \dots, \phi^{(K)}(\mathbf{x}) \right] \right).
\end{equation}

By training $\mathcal{M}_{g}$, it is capable of dynamically adjusting the contribution weights of each expert based on the input context. Notably, although the input dimension theoretically scales with $K$, thanks to the spatial mutual exclusivity derived from the CART tree partitioning, at most only the local expert covering the specific sample is activated for any given inference query. This sparse activation mechanism ensures that inference latency remains constant, perfectly aligning with strict industrial requirements for low latency.
\section{Experiments}

\begin{table*}[t]
    \centering
    \small 
    \setlength{\tabcolsep}{4pt}
    \begin{tabular}{l c c c c c c c c c c c}
        \toprule
        \textbf{Dataset} & \textbf{Base (XGB)} & \textbf{TabNet} & \textbf{FT-T.} & \textbf{AutoFeat} & \textbf{OpenFE} & \textbf{CAAFE} & \textbf{FeatLLM} & \textbf{OCTree} & \textbf{LLM-FE} & \textbf{Res-XGBoost} & \textbf{(Ours)} \\
        \midrule
        adult & 0.873 & 0.855 & 0.860 & $\times$ & 0.873 & 0.872 & 0.842 & 0.870 & \underline{0.874} & 0.874 & \textbf{0.875} \\
        bank-marketing & 0.906 & 0.905 & 0.908 & $\times$ & \underline{0.908} & 0.907 & 0.907 & 0.900 & 0.907 & 0.907 & \textbf{0.910} \\
        blood-transfusion & 0.742 & 0.756 & 0.777 & 0.738 & 0.747 & 0.749 & 0.771 & 0.755 & \underline{0.805} & 0.739 & \textbf{0.820} \\ 
        breast-w & 0.956 & 0.951 & 0.963 & 0.956 & 0.956 & 0.960 & 0.967 & 0.969 & \underline{0.970} & 0.960 & \textbf{0.986} \\
        credit-g & 0.751 & 0.693 & 0.740 & 0.757 & 0.758 & 0.751 & 0.707 & 0.753 & \underline{0.784} & 0.724 & \textbf{0.815} \\
        pc1 & 0.931 & 0.926 & 0.928 & 0.931 & 0.931 & 0.929 & 0.933 & 0.934 & \underline{0.935} & 0.928 & \textbf{0.946} \\  
        private-data$^\dagger$ & \underline{0.688} & 0.639 & 0.662 & 0.666 & 0.635 & $\times$ & $\times$ & $\times$ & $\times$ & 0.659 & \textbf{0.694} \\   
        \bottomrule
    \end{tabular}
    \caption{Performance comparison on various datasets. The evaluation metric is accuracy, except for the private dataset (marked with ``$\dagger$'') which is evaluated using AUC. Best results are highlighted in bold, and second-best are \underline{underlined}. Due to space constraints, detailed standard deviations are provided in Appendix Table~\ref{tab:std_dev_comparison}.}
    \label{tab:mean_performance}
\end{table*}

\subsection{Experimental Setup}
To conduct a comprehensive evaluation of our method, we construct a dataset benchmark consisting of two types of data with different characteristics in the experiment. First, to ensure reproducibility and alignment with the SOTA methods, we select six widely utilized public datasets from OpenML~\cite{feurer2021openml,vanschoren2014openml}. These datasets cover multiple high-stakes domains, with sample sizes ranging from 699 to 48,000. Second, to further validate the model's deployment capability in real-world industrial scenarios, we incorporated a large-scale private dataset for financial risk control from Qfin Holdings (private-data). Derived from actual business workflows, this dataset is characterized by high feature dimensionality and significant noise, presenting a level of complexity that far exceeds that of standard public benchmarks.

For this private dataset, we implement two necessary adjustments to our experimental settings. First,  due to strict data privacy compliance requirements, sensitive data cannot be uploaded to the cloud. Therefore, LLM methods relying on external APIs, e.g., CAAFE, FeatLLM, are not evaluated on this dataset. Second, given the severe class imbalance prevalent in risk control scenarios, we employ AUC as the primary evaluation metric for this dataset, while retaining accuracy for the remaining public datasets. Regarding the general experimental protocol, we uniformly adopt an 80\% training set and 20\% test set division for all datasets and, following the standard protocol of~\cite{hollmann2023large}, conduct evaluations over five random seeds, reporting the final results as averages. For further details, please refer to Appendix \ref{apx:dataset}.

We compare our approach against the most widely used models for tabular data, TabNet~\cite{arik2021tabnet} and FT-Transformer~\cite{gorishniy2021revisiting}, as well as SOTA feature engineering techniques, including OpenFE~\cite{zhang2023openfe} and AutoFeat~\cite{horn2019autofeat}. We also include LLM-based methods such as CAAFE~\cite{hollmann2023large}, FeatLLM~\cite{han2024large}, OCTree~\cite{nam2024optimized}, and LLM-FE~\cite{abhyankar2025llm}. For the feature engineering comparisons, we utilize XGBoost as the default prediction model for tabular data and performed 30 Bayesian hyperparameter optimizations on it. We use GPT-3.5-Turbo as the standard LLM infrastructure for all LLM-based benchmark models. To ensure a fair comparison, all LLM-based baselines are configured to query the LLM backbone for a total of 20 samples, iterating until optimal performance is achieved. Appendix~\ref{apx:baseline} contains additional implementation details.

In our experiments, we employ the locally deployed Seed-OSS-36B-Instruct~\cite{seed2025seed-oss} to ensure strict data compliance while leveraging raw feature semantics. We set the sampling temperature to $t = 0.1$ and the number of experts to $K = 5$. To ensure consistency with baseline methods, NSR-Boost supplies 20 data samples to the LLM during each iteration and terminates after 5 successful iterations, designating the resulting function as the final expert function.

\subsection{Offline Performance Analysis}

\begin{figure}[t]
    \includegraphics[width=\linewidth]{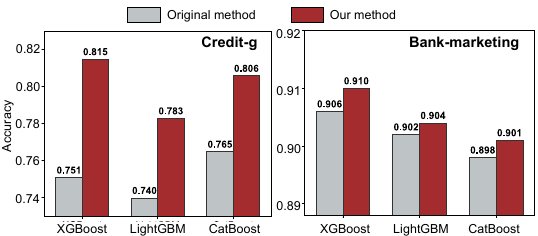}  
    \caption{Generalizability analysis of NSR-Boost across various backbones and datasets.}
    \label{fig:normal}
\end{figure}
\textbf{Comparison of all methods. }As shown in Table~\ref{tab:mean_performance}, NSR-Boost exhibits superior performance across six public datasets from diverse domains, achieving a mean rank of $1.0$. Specifically, our framework demonstrates exceptional adaptability across varying data scales. On data-scarce datasets such as credit-g and blood-transfusion, the semantic priors injected by the LLM serve as effective regularization against noise, boosting XGBoost's accuracy by $+6.4\%$ and $+7.8\%$, respectively. On large-scale benchmarks like adult and bank-marketing, where GBDTs are notoriously difficult to surpass, NSR-Boost still extracts additional performance gains by capturing complex non-linear interactions within the long tail of the distribution. Furthermore, on the real-world private dataset, we observe an even more significant advantage in noise robustness. Specifically, the simple residual boosting baseline suffered a performance regression due to overfitting the high-frequency noise inherent in industrial data, with the AUC plummeting from $0.688$ to $0.659$. In contrast, by leveraging symbolic experts, NSR-Boost successfully filtered out random noise and extracted robust incremental signals, robustly elevating the AUC to $0.694$.

\noindent \textbf{Generalization Capability of NSR-Boost.} The operation of NSR-Boost does not interfere with the execution or deployment of existing models, which gives it extremely high flexibility in industrial applications. Figure~\ref{fig:normal} illustrates the performance gains achieved when applying NSR-Boost across diverse gradient boosting backbones. Experimental results demonstrate that, regardless of variations in the underlying base model, NSR-Boost consistently enhances prediction accuracy. For example, on the credit-g dataset, the performance of LightGBM is improved by +4.3\%. This finding confirms that the symbolic residuals mined by our framework are not merely overfitting to the specific biases of a particular algorithm. Instead, they successfully capture the intrinsic properties of the data. Consequently, this positions NSR-Boost as a universal performance enhancement solution capable of seamlessly adapting to various legacy systems.

\subsection{Interpretability and Ablation Study}
\begin{figure}[t]
    \includegraphics[width=\linewidth]{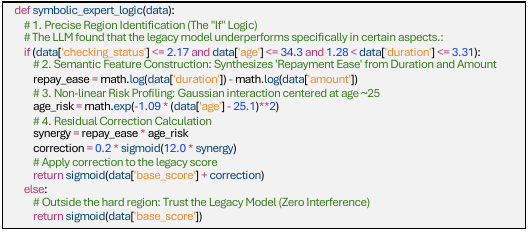}  
    \caption{The interpretable logic of a Symbolic Expert.}
    \label{fig:casecode}
\end{figure}
Figure~\ref{fig:casecode} illustrates the generated symbolic expert code generated on the credit-g dataset and intuitively demonstrates the white-box interpretability and safety mechanisms of NSR-Boost. The code employs explicit if-else logic structures to precisely delineate the operational boundaries of residual correction. The model activates and computes additional correction terms only when samples fall within specific regions. It directly outputs the original predictions of the legacy model in all other instances to ensure zero interference with non-target groups. Furthermore the automatically constructed features possess clear business semantics that enable business personnel to directly review and comprehend the decision logic. This implementation thereby achieves a unification of high performance and high transparency. A more detailed analysis is provided in Appendix~\ref{apx:inter}.

\begin{figure}[t]
    \includegraphics[width=\linewidth]{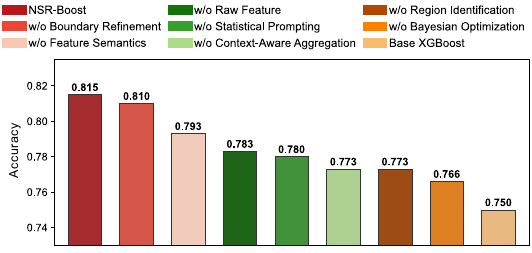}  
    \caption{Ablation study of NSR-Boost components on the credit-g dataset.}
    \label{fig:ablation}
\end{figure}

Figure~\ref{fig:ablation} presents the ablation study results of NSR-Boost on the credit-g dataset, verifying the necessity of each module. Firstly, Bayesian optimization holds a central position in the framework. Removing this component has the greatest impact on performance, causing the accuracy to drop from $0.815$ to $0.766$, which clearly demonstrates that although LLM excels at generating symbolic code structures, it is difficult to directly determine precise continuous parameters. It must rely on a hybrid optimization strategy to make up for the shortcomings in numerical accuracy. Secondly, the absence of context-aware aggregation and region identification both lead to an accuracy drop to $0.773$. This confirms that merely generating the correct residual function is not enough. It must be combined with the identification of hard regions and a robust dynamic fusion mechanism to avoid the noise caused by forced fitting by expert models in the entire sample space.

Regarding information utilization and context dependency, experimental results reveal the distinct value of different information modalities. First, the performance decline of $0.035$ caused by removing statistical prompting proves that statistical metrics are important anchors for the LLM to rapidly locate high-value features. Meanwhile, the removal of feature semantics similarly causes accuracy loss, indicating that natural language descriptions of features provide necessary supplementary context for the model to understand business meanings. Second, when we remove raw feature inputs at the final aggregation stage, model performance dropped to $0.783$. This indicates that the aggregator does not perform blind weight allocation, but rather relies on the complete context of the raw feature space to precisely perceive the performance discrepancies of the baseline model on specific instances, thereby achieving the precise invocation of symbolic experts.

Finally, although boundary refinement yields a smaller absolute improvement, this $0.5\%$ gap precisely demonstrates that fine-tuning samples near the decision boundary can further unlock the model's potential. In summary, the superiority of NSR-Boost stems from the synergy of each component, and even the weakest variant still outperforms the original XGBoost, which proves the inherent robustness of the framework.
\subsection{Online Deployment}
\begin{figure}[t]
    \includegraphics[width=\linewidth]{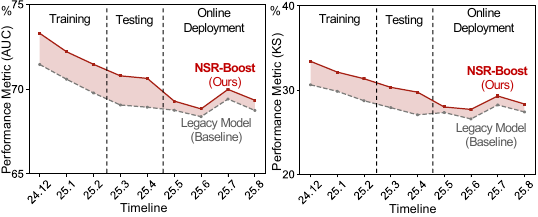}  
    \caption{Performance evaluation of the Emerging Customer Segment (B-Card) model. The online deployment processes approximately 2.4 million real user samples.}
    \label{fig:online}
\end{figure}

\begin{figure}[t]
    \includegraphics[width=\linewidth]{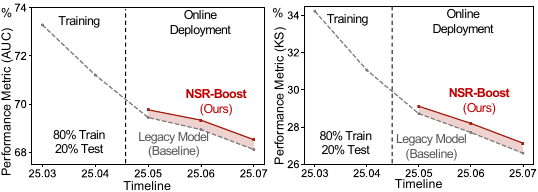}  
    \caption{Performance evaluation of the Portfolio Management Segment (B-Card) model. The online deployment processes approximately 1.1 million real user samples.}
    \label{fig:online2}
\end{figure}
To support the routine application of NSR-Boost in assets with a scale of hundreds of billions, Qfin Holdings has established a high-performance computing infrastructure named ``NSR-Platform'', and conducted a comprehensive online evaluation on this basis. 

In terms of Infrastructure and Energy Efficiency, the platform's underlying computational power is supported by a high-performance computing cluster comprising 2 groups totaling 16 H20 accelerator cards. This infrastructure delivers the throughput capable of processing 8 core business models daily, equivalent to the parallel generation of 96 chains. Given that the chain construction process involves complex Bayesian optimization and XGBoost model building, which are typically CPU-intensive tasks, we allocate dedicated, high-specification CPU resources to the generation of each computational chain: a 12-core CPU [Intel(R) Xeon(R) Silver 4214 @ 2.20GHz] paired with 60GB of memory.

To achieve ultimate cost reduction, the platform adopts an elastic computing strategy. Computational instances are dynamically instantiated only upon task triggering and are immediately released upon completion. Calculations indicate that generating a full suite of 12 computational chains for a single business model consumes approximately 48 GPU-hours ($16 \text{GPUs} \times 24\text{h} / 8 \text{ models}$). Relative to the business gains generated by the model on a scale of hundreds of billions (RMB), this computational cost is virtually negligible.

NSR-Platform has cumulatively optimized business models approximately 130 times, extensively covering core scenarios such as Risk Control, Fund Drawdown, and User Login. Statistical data indicates that models optimized by NSR-Platform achieve average improvements exceeding 0.3\% in both AUC and Kolmogorov-Smirnov (KS) metrics. Given the platform's massive financial throughput, a robust metric enhancement of 0.3\% to 0.5\% translates into significant potential savings in bad debt reduction.

Next, we report the online deployment results of NSR-Boost in two typical scenarios, each processing millions of real-world samples. Since NSR-Boost operates as an additional enhancement to the legacy model, the prediction results of both the legacy model and NSR-Boost are kept and recorded during the online deployment process, enabling a direct comparison of the two on the same input request without traffic splitting.

Figure~\ref{fig:online} illustrates the real-world online performance within the ``Emerging Customer Segment (B-Card)'' scenario. It is noteworthy that the legacy model in this context is not a weak baseline, but a full-scale XGBoost model finely tuned on 300-400 expert features. Even atop such a high-dimensional and robust baseline, NSR-Boost demonstrates superior advantages. During both the offline test period and the strictly out-of-time (OOT) online deployment phase, NSR-Boost consistently dominates the baseline, achieving an average AUC improvement of 1.17\% and an average KS improvement of 1.81\%. Crucially, this technical advantage translates directly into substantial business value. By leveraging NSR-Boost scores to prioritize high-quality users, the bad rate is significantly reduced from 3.04\% to 2.86\%, proving that the model successfully captures risk patterns overlooked by the legacy model.

Similarly, Figure~\ref{fig:online2} depicts performance in the ``Portfolio Management Segment (B-Card)'' scenario, with a particular emphasis on the rigorous OOT deployment phase. Here, the legacy model remains a robust XGBoost model. Despite its strength, the baseline suffered performance degradation during the OOT phase due to concept drift. In contrast, NSR-Boost delivered consistent gains, realizing an AUC improvement of 0.37\% and a KS improvement of 0.47\%, and compresses the defect rate from 2.96\% to 2.81\%.

\subsection{Efficiency and Hyperparameter Analysis}
\begin{figure}[t]
    \includegraphics[width=\linewidth]{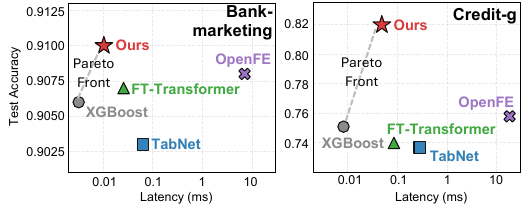}  
    \caption{Inference latency vs. accuracy Pareto frontier for different methods.}
    \label{fig:efficiency}
\end{figure}

\textbf{Inference Latency and Pareto Frontier Analysis.} Industrial applications impose stringent requirements on response times for online services, typically mandating that single-inference latency be controlled within the millisecond range to support high-concurrency demands. To evaluate the deployment feasibility of NSR-Boost, we compare the trade-off between test accuracy and inference latency across different methods in Figure~\ref{fig:efficiency}. As illustrated, although deep tabular models such as FT-Transformer and TabNet exhibit competitive accuracy, their inference latencies typically reach the order of 10ms or even 100ms. This prohibitive computational cost not only necessitates expensive GPU clusters but also fails to meet the rigid low-latency benchmarks required for real-time risk control scenarios. In contrast, NSR-Boost is positioned on the Pareto Frontier of the accuracy-latency trade-off. By leveraging its generated pure Python code format, NSR-Boost maintains SOTA accuracy while constraining inference latency to under 1ms. Compared to OpenFE, we achieve higher accuracy at equivalent latency levels. Compared to XGBoost, we realize a significant leap in accuracy with only a negligible increase in computational overhead (at the microsecond level). This demonstrates that NSR-Boost is currently the unique solution capable of simultaneously satisfying the dual criteria of high precision and low latency, enabling seamless deployment without requiring upgrades to existing hardware infrastructure.
\begin{figure}[t]
    \includegraphics[width=\linewidth]{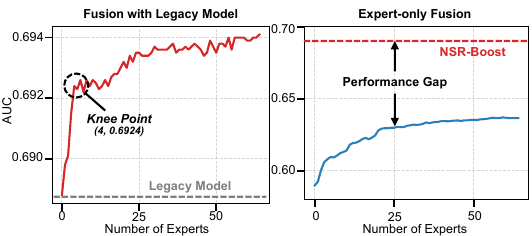}  
    \caption{Impact of the number of Symbolic Experts on the industrial financial private dataset.}
    \label{fig:convergence}
\end{figure}

\noindent\textbf{Hyperparameter Sensitivity and Residual Mechanism Verification.} To identify the optimal balance between inference precision and system load for industrial deployment, we conducted a sensitivity analysis on the number of Symbolic Experts ($K$) using a large-scale private dataset. As illustrated in the left panel of Figure~\ref{fig:convergence}, the AUC of the hybrid model exhibits a typical trend of diminishing returns as the number of experts increases. The curve displays a distinct ``Knee Point'' at $K=4$: At this juncture, the model has already secured the vast majority of performance gains. Continuing to increase the number of experts yields only negligible improvements in accuracy while linearly increasing inference latency.

Furthermore, the experimental results in the right panel of Figure~\ref{fig:convergence} vigorously refute the conventional notion that a new model must inherently outperform the old one, profoundly revealing the residual correction nature of NSR-Boost. Experiments indicate that if the legacy model is detached and prediction relies solely on the aggregated Symbolic Experts (Expert-only Fusion), the AUC reaches only approximately 0.64, which is significantly lower than the legacy model baseline. However, it is precisely the combination of these weak experts with the legacy model that propels overall performance to SOTA levels. This phenomenon confirms that NSR-Boost does not attempt to relearn global patterns from scratch, but rather specifically targets and fills the blind spots of the legacy model regarding long-tail samples.

\section{Conclusion}
We propose NSR-Boost, a non-intrusive framework that enhances the performance of frozen legacy models by repairing ``hard regions'' via neuro-symbolic residual learning. Our approach effectively bridges the gap between the discrete logic generation of LLMs and the continuous parameter optimization required for high-precision tasks. Empirical results demonstrate that NSR-Boost not only achieves SOTA performance against strong baselines but also uncovers meaningful white-box business logic. This work offers a practical, safe, and interpretable pathway for the continuous evolution of large-scale industrial systems.

\bibliographystyle{ACM-Reference-Format}
\bibliography{sample-base}

\appendix

\section{Appendix}

\subsection{LLM Usage Statement}
During the preparation of this manuscript, the authors utilized an LLM exclusively for grammar and language polishing. All technical content, scientific claims, and experimental results were conceived, derived, and verified solely by the authors.

\subsection{Dataset Details}
\label{apx:dataset}

Table~\ref{tab:dataset} details the collection of binary classification datasets utilized for evaluation, primarily sourced from OpenML, alongside a private dataset from Qfin Holdings. Given that the high-stakes domains discussed in this paper are typically modeled as binary classification problems, we place particular emphasis on such tasks. In order to meet the interpretability requirements of the business, we deliberately choose those datasets with descriptive feature names. Each dataset is accompanied by a task description to facilitate an understanding of the background context. Our selection spans from small-scale to large-scale datasets. This diverse setting, characterized by varying feature dimensionalities and sample sizes, represents a wide spectrum of real-world scenarios, thereby providing a robust framework for evaluating the proposed method.
\begin{table}[htbp]
    \centering
    \small 
    \begin{tabular}{l c c c c}
        \toprule
        \textbf{Dataset} & \textbf{Features} & \textbf{Samples} & \textbf{Source} & \textbf{ID}\\
        \midrule
        adult & 14 & 48842 & OpenML & 1590 \\
        blood-transfusion & 4 & 748 & OpenML & 1464 \\
        bank-marketing & 16 & 45211 & OpenML & 1461 \\
        breast-w & 9 & 699 & OpenML & 15 \\
        credit-g & 20 & 1000 & OpenML & 31 \\
        pc1 & 21 & 1109 & OpenML & 1068\\
        private-data & 200 & 497677 & Qfin Holdings & -\\
        \bottomrule
    \end{tabular}
    \caption{Dataset statistics.}
    \label{tab:dataset}
\end{table}

\subsection{Baseline Details}
\label{apx:baseline}

\begin{figure*}[t]
    \includegraphics[width=\textwidth]{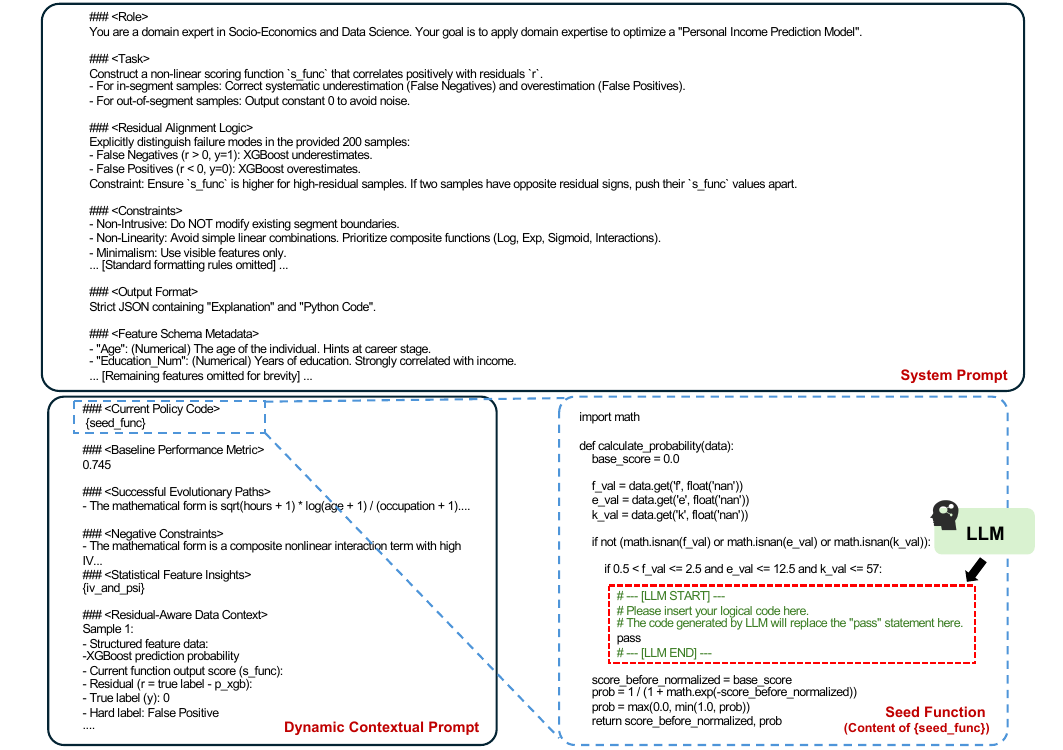}  
    \caption{The prompt engineering framework for Symbolic Expert generation.}
    \label{fig:prompt}
\end{figure*}

\begin{figure*}[t]
    \includegraphics[width=\textwidth]{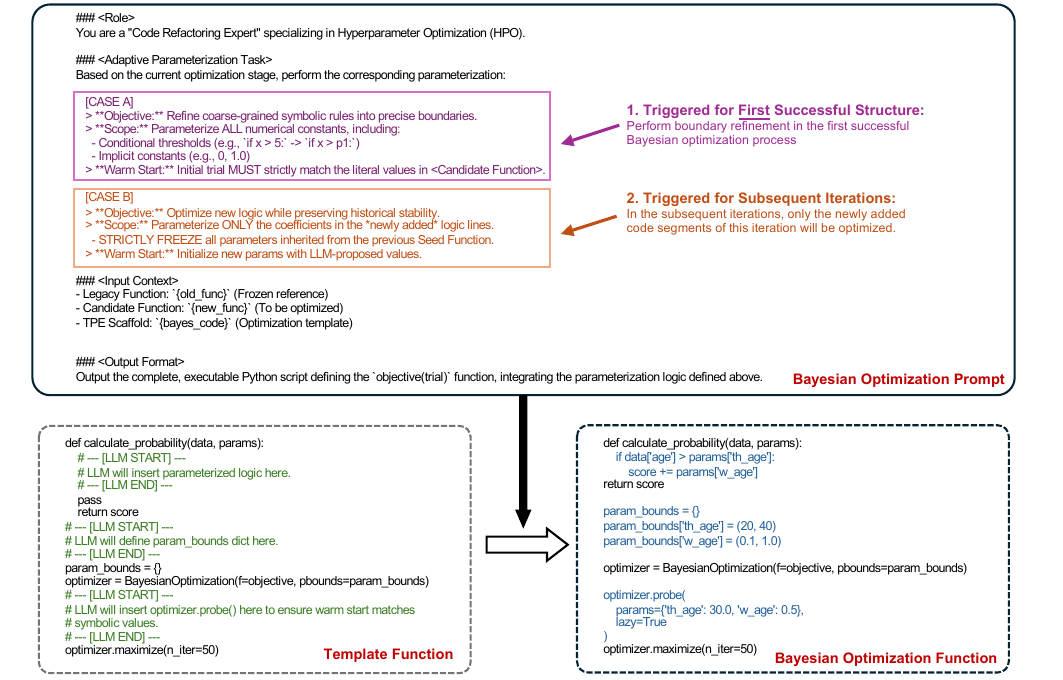} 
    \caption{The prompt engineering framework for Bayesian optimization.}
    \label{fig:bo_prompt}
\end{figure*}

To fully evaluate the efficacy of NSR-Boost, we compare it against a wide range of baselines, spanning from traditional GBDT and deep tabular models to recent LLM-based feature engineering methods. To ensure a fair comparison, we enforce a unified data preprocessing pipeline and consistent evaluation protocols across all experiments. Specifically, we implement the following baselines:
\begin{itemize}
    \item \textbf{TabNet} employs a sequential attention mechanism to realize interpretable instance-wise feature selection and inference. We implement it using its official library and maintained the default hyperparameter settings.
    \item \textbf{FT-Transformer} successfully adapts deep attention architectures to tabular data by transforming features into embeddings and feeding them into Transformer layers. We use the official implementation of this method and adopt the default configuration to ensure a fair comparison.
    \item \textbf{AutoFeat} combines iterative feature subsampling with beam search strategies, aiming to efficiently generate non-linear features. In our experiments, we implemented this method based on the official open-source library \texttt{autofeat} and maintained its default hyperparameter configuration.
    \item \textbf{OpenFE} employs feature boosting and stepwise pruning strategies to screen for effective high-order interaction features from a vast candidate space. We use its official library \texttt{openfe} directly and adhere to the recommended standard configuration for our experiments.
    \item \textbf{CAAFE} exploits the semantic understanding capabilities of LLMs to generate features with explicit business significance. We implement this baseline based on its official library, strictly adhering to its original hyperparameter settings and standard preprocessing workflow, and fed the augmented data into the downstream model.
    \item \textbf{FeatLLM} leverages LLMs to generate explicit rules for feature binarization.  To construct a stronger baseline, we upgrade the downstream linear model to XGBoost and adopt an asymmetric data strategy. It samples only 10 instances for rule generation while uses the full training set for parameter fitting. Final results are averaged over three independent runs to mitigate the stochasticity associated with few-shot learning.
    \item \textbf{OCTree.} For OCTree, we use its official implementation, making only minor adjustments to the data loading and model initialization modules to adapt to the unified evaluation framework of this experiment.
    \item \textbf{LLM-FE} is another automated feature engineering approach driven by LLMs. In our experiments, we strictly follow the standard configuration outlined in its original paper and adopt the default hyperparameter settings to ensure reproduction.
    
    \item \textbf{Res-XGBoost} is a residual learning baseline constructed to verify the necessity of symbolic regression.  This method directly uses gradient boosting trees to fit the residuals of the base model. In the experiments, we introduce a Bayesian optimization strategy to automatically adjust the learning rate, tree depth, and regularization term.
\end{itemize}

\subsection{NSR-Boost Prompt Details}
\label{apx:prompt}
Figure~\ref{fig:prompt} illustrates the Residual-Driven Symbolic Generation process using the adult dataset, where the LLM acts as a socio-economic expert to generate correction functions based on dynamic residual feedback. Figure~\ref{fig:bo_prompt} details the Bayesian Optimization, showcasing the Adaptive Parameterization Task where the LLM distinguishes between Boundary Refinement (CASE A) and Incremental Search (CASE B) to refactor symbolic logic into a numerical optimization template with warm-start constraints.

\section{Additional Results}

\begin{table*}[t]
    \centering
    \small 
    \setlength{\tabcolsep}{4pt}
    \begin{tabular}{l c c c c c c c c c c c}
        \toprule
        \textbf{Dataset} & \textbf{Base (XGB)} & \textbf{TabNet} & \textbf{FT-T.} & \textbf{AutoFeat} & \textbf{OpenFE} & \textbf{CAAFE} & \textbf{FeatLLM} & \textbf{OCTree} & \textbf{LLM-FE} & \textbf{Res-XGBoost} & \textbf{(Ours)} \\
        \midrule
        adult & $\pm 0.002$ & $\pm 0.005$ & $\pm 0.002$ & $\times$ & $\pm 0.002$ & $\pm 0.002$ & $\pm 0.003$ & $\pm 0.002$ & $\pm 0.003$ & $\pm 0.001$ & $\pm 0.002$ \\
        bank-marketing & $\pm 0.003$ & $\pm 0.002$ & $\pm 0.001$ & $\times$ & $\pm 0.002$ & $\pm 0.002$ & $\pm 0.002$ & $\pm 0.002$ & $\pm 0.002$ & $\pm 0.001$ & $\pm 0.002$ \\
        blood-transfusion & $\pm 0.012$ & $\pm 0.033$ & $\pm 0.032$ & $\pm 0.014$ & $\pm 0.025$ & $\pm 0.017$ & $\pm 0.016$ & $\pm 0.026$ & $\pm 0.036$ & $\pm 0.002$ & $\pm 0.040$ \\
        breast-w & $\pm 0.012$ & $\pm 0.018$ & $\pm 0.006$ & $\pm 0.019$ & $\pm 0.014$ & $\pm 0.009$ & $\pm 0.015$ & $\pm 0.009$ & $\pm 0.009$ & $\pm 0.005$ & $\pm 0.010$ \\
        credit-g & $\pm 0.019$ & $\pm 0.020$ & $\pm 0.051$ & $\pm 0.017$ & $\pm 0.017$ & $\pm 0.020$ & $\pm 0.034$ & $\pm 0.021$ & $\pm 0.015$ & $\pm 0.004$ & $\pm 0.017$ \\
        pc1 & $\pm 0.004$ & $\pm 0.008$ & $\pm 0.006$ & $\pm 0.014$ & $\pm 0.009$ & $\pm 0.005$ & $\pm 0.007$ & $\pm 0.007$ & $\pm 0.006$ & $\pm 0.003$ & $\pm 0.006$ \\    
        \bottomrule
    \end{tabular}
    \caption{Standard deviation of performance on various datasets over five independent runs.}
    \label{tab:std_dev_comparison}
\end{table*}

\subsection{Experimental Stability Analysis} 
To evaluate the robustness of NSR-Boost against the stochasticity inherent in LLM generation and optimization processes, we report the performance standard deviations over five independent runs in Table~\ref{tab:std_dev_comparison}. The results indicate that NSR-Boost maintains low variance levels, which are comparable to deterministic numerical baselines.

\begin{figure}[t]
    \includegraphics[width=\linewidth]{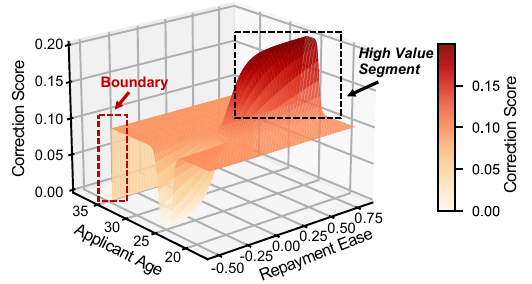}  
    \caption{Visualization of the Symbolic Expert's decision surface in Figure~\ref{fig:casecode}.}
    \label{fig:casestudy}
\end{figure}
\subsection{Interpretability}
\label{apx:inter}
Figure~\ref{fig:casestudy} provides a geometric visualization of the symbolic expert logic presented in Figure~\ref{fig:casecode} of the main text. This three dimensional surface bridges abstract mathematical formulas with their concrete behaviors within the decision space. The steep vertical truncation displayed within the red dashed box directly maps to the conditional statements found in the source code. This distinct visual feature provides compelling evidence that the model strictly enforces safety constraints by completely halting the correction mechanism for any applicants older than 34.3 years. Furthermore the smooth peak like elevation enclosed in the black dashed box offers an intuitive visual representation of the Gaussian interaction term defined within the algorithm. This geometric form confirms that the Symbolic Expert successfully identified a nonlinear opportunity window centered at 25.1 years wherein applicants possessing high repayment slack are recognized as high quality candidates and receive positive score corrections.

\subsection{Performance on High-Value Customer Segment}

To further verify the generalization capability of NSR-Boost across distinct business lines, we report supplementary deployment results from a third core industrial scenario targeting the High Value Segment of the B Card.

Table~\ref{tab:detailed_performance} details the monitoring data derived from processing approximately 1.41 million real world samples between February 2025 and July 2025. Consistent with the primary experimental setup, we maintained the Legacy Model in a frozen state. Experimental results demonstrate that NSR-Boost achieved sustained performance improvements within this scenario. Notably, NSR-Boost achieves a weighted average AUC improvement of 0.43\% and a weighted average KS improvement of 0.92\% during the strict OOT online deployment phase from May 2025 to July 2025.
\begin{table*}[t]
    \small
    \centering
    \setlength{\tabcolsep}{5pt} 
    \begin{tabular}{llcccccc}
        \toprule
        \multirow{2}{*}{\textbf{Phase}} & \multirow{2}{*}{\textbf{Month}} & \multicolumn{2}{c}{\textbf{Legacy Model}} & \multicolumn{4}{c}{\textbf{NSR-Boost}} \\
        \cmidrule(lr){3-4} \cmidrule(lr){5-8}
         & & \textbf{AUC} & \textbf{KS} & \textbf{AUC} & \textbf{KS} & \textbf{$\Delta$ AUC} & \textbf{$\Delta$ KS} \\
        \midrule
        \multirow{3}{*}{Training} 
         & Feb-25 & 0.7542 & 0.3716 & 0.7685 & 0.3940 & 0.0142$\uparrow$ & 0.0224$\uparrow$ \\
         & Mar-25 & 0.7456 & 0.3595 & 0.7631 & 0.3871 & 0.0175$\uparrow$ & 0.0276$\uparrow$ \\
         & Apr-25 & 0.7405 & 0.3576 & 0.7551 & 0.3801 & 0.0146$\uparrow$ & 0.0225$\uparrow$ \\
        \midrule
        \multirow{4}{*}{\textbf{Online Deployment}} 
         & May-25 & 0.7477 & 0.3644 & 0.7532 & 0.3723 & \textbf{0.0055$\uparrow$} & \textbf{0.0079$\uparrow$} \\
         & Jun-25 & 0.7443 & 0.3549 & 0.7471 & 0.3636 & \textbf{0.0027$\uparrow$} & \textbf{0.0087$\uparrow$} \\
         & Jul-25 & 0.7341 & 0.3388 & 0.7389 & 0.3499 & \textbf{0.0047$\uparrow$} & \textbf{0.0111$\uparrow$} \\
         \cmidrule(lr){2-8}
         & Avg (May-Jul) & 0.7420 & 0.3527 & 0.7464 & 0.3619 & \textbf{0.0043$\uparrow$} & \textbf{0.0092$\uparrow$} \\
        \bottomrule
        \multicolumn{8}{l}{\footnotesize \textit{Note:} The training data (Feb-25 to Apr-25) contains mixed validation samples, shuffled and split with an 80\%/20\% ratio.} \\
    \end{tabular}
    \caption{Performance evaluation of the High-Value Segment (B-Card) model. The online deployment processes approximately 1.41 million real user samples.}
    \label{tab:detailed_performance}
\end{table*}

\subsection{LLM Model Selection and Stability Analysis} 
\begin{figure}[t]
    \includegraphics[width=\linewidth]{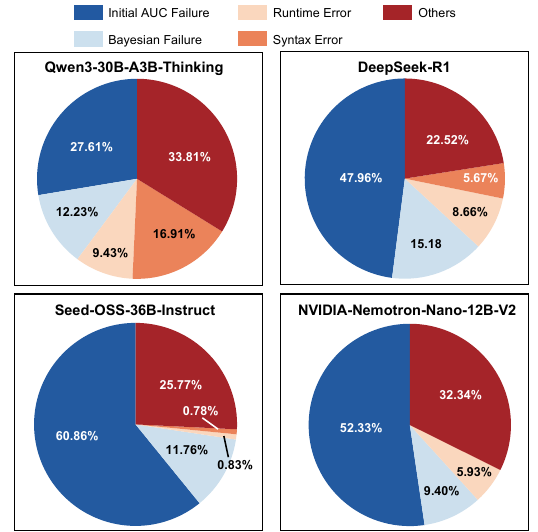}  
    \caption{Distribution of failure modes during the symbolic expert generation phase across four different LLMs.}
    \label{fig:pie}
\end{figure}

As illustrated in Figure~\ref{fig:pie}, we conduct a comprehensive analysis of the failure modes of four different LLMs, namely DeepSeek-R1~\cite{deepseekai2025deepseekr1incentivizingreasoningcapability}, Qwen3-30B-A3B-Thinking~\cite{qwen3technicalreport}, NVIDIA-Nemotron-Nano-12B-V2~\cite{nvidia2025nvidianemotronnano2}, and Seed-OSS-36B-Instruct~\cite{seed2025seed-oss},reveals a consistent pattern: The primary bottleneck in expert code construction lies not in syntactic correctness, but in semantic validity. The dominant failure category, ``Initial AUC Failure'', accounts for a substantial proportion of all rejected cases, ranging from approximately 27\% to over 60\% depending on the model. This phenomenon manifests as a form of hallucination where the LLM-generated Python code is syntactically flawless and executable yet fails to capture the underlying residual patterns, resulting in no substantive improvement in validation AUC.

In contrast, explicit errors such as Runtime Error and Syntax Error remain at remarkably low levels, particularly for models with strong instruction-following capabilities like Seed-OSS-36B and DeepSeek-R1. This highlights that while modern LLMs have evolved into proficient programmers capable of strictly adhering to formatting constraints, they remain unreliable data scientists when performing unsupervised reasoning tasks. Therefore, these findings provide strong empirical evidence for the rigorous validation process adopted in NSR-Boost.

\subsection{Generalization Ability across LLMs.}
Based on this rigorous verification mechanism, we further explore whether the effectiveness of NSR-Boost is bound to a specific LLM. We conduct a sensitivity analysis using the four aforementioned open-source LLMs, executing 20 independent runs for each model on the private financial dataset.

As shown in Table~\ref{tab:llm_backbone}, all four backbones successfully enable the framework to outperform the initial baseline. It is worth noting that the Seed-OSS-36B model and DeepSeek-R1, which exhibit strong inference ability, achieve the top two results. The extremely small performance gap between different architectures proves that NSR-Boost is model-agnostic and robust, and can extract stable incremental gains regardless of the LLM. Based on these empirical results, especially considering the maximum improvement in performance, we ultimately choose Seed-OSS-36B.
\begin{table}[htbp]
\centering

\begin{tabular}{l|c|c}
\toprule
\textbf{LLM Model} & \textbf{Average AUC} & \textbf{Improvement} \\
\midrule
\textit{Legacy Model} & 0.688982 & - \\
\midrule
Nemotron-Nano-12B & 0.689159 & +0.000177 \\
Qwen3-30B & 0.689249 & +0.000267 \\
DeepSeek-R1 & 0.689302 & +0.000320 \\
\textbf{Seed-OSS-36B (Ours)} & \textbf{0.689388} & \textbf{+0.000406} \\
\bottomrule
\end{tabular}
\caption{Performance comparison of different LLMs on the private dataset.}
\label{tab:llm_backbone}
\end{table}

\subsection{In-depth Analysis of Residual Correction Benchmarks.} 

\begin{table}[t]
\centering
\begin{tabular}{lccc}
\toprule
\textbf{Method} & \textbf{Paradigm} & \textbf{AUC Gain} & \textbf{Interp.} \\ \midrule
\textit{Legacy Model} & GBDT & - & High \\
Res-XGBoost & Black-box & -1.02e-3 & Low \\ \midrule
gplearn (20min) & SR & -1.78e-3 & Moderate \\
gplearn (8.3h) & SR & $\pm$0.00e-3 & Low \\ \midrule
\textbf{NSR-Boost} & \textbf{Neuro-Symbolic} & \textbf{+4.06e-4} & \textbf{High} \\ \bottomrule
\end{tabular}
\caption{Performance and interpretability comparison of residual learning methods on the private OOT test set.}
\label{tab:sr_comparison}
\end{table}

Table~\ref{tab:sr_comparison} presents a detailed comparison of residual correction performance across different paradigms on the OOT test set of the private dataset.

Purely data-driven baseline methods encountered significant generalization bottlenecks. As a representative black-box model, Res-XGBoost exhibited a negative gain of $-1.02\text{e-}3$ despite undergoing extensive hyperparameter optimization via Optuna. This outcome indicates that given the robustness of the Legacy Model, the remaining residuals contain substantial random noise. Consequently, unconstrained strong estimators are prone to overfitting this noise, leading to a marked degradation in performance on the test set.

Traditional symbolic regression methods exposed their limitations regarding search efficiency in high-dimensional spaces. Under the standard setting of a 20-minute search, gplearn similarly succumbed to overfitting. More instructively, even after investing substantial computational resources for an intensive 8.3-hour search, the algorithm ultimately converged to a gain of $\pm 0.00$ with its final output being a trivial constant solution of -0.01. This phenomenon demonstrates that without the guidance of prior knowledge, genetic algorithms struggle to overcome the failure of search strategies within complex operator spaces caused by high-dimensional feature combinations.

In contrast, NSR-Boost stands out as the sole method in the experiment to achieve a positive gain of $+4.06\text{e-}4$. By leveraging semantic understanding to effectively prune the search space, NSR-Boost enables the model to precisely locate feature interactions with clear business logic within extremely low signal-to-noise environments. This capability allows it to circumvent overfitting while simultaneously breaking through the performance ceiling inherent to purely data-driven methods.

Furthermore, regarding interpretability, although Res-XGBoost possesses strong fitting capabilities, its nature as a black-box ensemble composed of thousands of trees results in a total loss of physical interpretability. Meanwhile, Symbolic Regression in high search spaces tends to exhibit two extremes, yielding either unreadable long formulas due to data complexity or degenerating into trivial solutions to pursue generalization as shown in the table. Conversely, NSR-Boost strikes a balance between these extremes, achieving a unification of interpretability and performance.

\subsection{Aggregator Behavior and Baseline Fairness}

We further analyze whether the final prediction is dominated by the XGBoost aggregator. Global SHAP values show that the dominant factors are expert-base interaction terms and the legacy score: Expert-Base Product (0.33), Legacy Base Score (0.32), \texttt{raw\_age} (0.26), and Expert-Base Difference (0.23). Most raw features have zero SHAP impact, indicating that the aggregator primarily acts as a router and fallback controller rather than an unconstrained black-box predictor.

\begin{table}[t]
\centering
\setlength{\tabcolsep}{5pt}
\begin{tabular}{lc}
\toprule
\textbf{Dataset} & \textbf{$\Delta$ Acc. (Seed-OSS - GPT-3.5)} \\
\midrule
adult & 0.000 \\
bank-marketing & -0.004 \\
blood-transfusion & -0.002 \\
breast-w & +0.011 \\
credit-g & -0.031 \\
pc1 & +0.008 \\
\bottomrule
\end{tabular}
\caption{Accuracy differences of CAAFE when replacing GPT-3.5-Turbo with the locally deployed Seed-OSS backbone.}
\label{tab:caafe_seedoss}
\end{table}

We also reran CAAFE with the same locally deployed Seed-OSS backbone used by NSR-Boost. As shown in Table~\ref{tab:caafe_seedoss}, Seed-OSS does not yield systematic improvements over GPT-3.5-Turbo for CAAFE, suggesting that NSR-Boost's gains are attributable to the neuro-symbolic residual architecture rather than merely to a stronger LLM backbone.

\subsection{Statistical Significance}

We conduct significance tests over the public datasets. NSR-Boost achieves statistically significant improvements with p-values below 0.05 on all evaluated datasets: adult (0.0118), bank-marketing (0.0059), blood-transfusion (0.0088), breast-w (0.0124), credit-g (0.0056), and pc1 (0.0098).

\end{document}